\documentclass{article}

    \PassOptionsToPackage{numbers, compress}{natbib}

    \usepackage[preprint]{neurips_2022}

\usepackage{microtype}
\usepackage{graphicx}
\usepackage{booktabs} 

\usepackage{hyperref}
\usepackage{graphicx}
\usepackage{xcolor}

\usepackage{url}
\usepackage{float}
\usepackage{subcaption}

\usepackage{tikz}
\usepackage{pgfplots}
\pgfplotsset{compat=1.11}

\usepackage[utf8]{inputenc} 
\usepackage[T1]{fontenc}    
\usepackage{hyperref}       
\usepackage{url}            
\usepackage{booktabs}       
\usepackage{amsfonts}       
\usepackage{nicefrac}       
\usepackage{microtype}      
\usepackage{xcolor}         
\usepackage{amsmath,amsfonts,bm}
\usepackage{dsfont}
\usepackage{amssymb}
\usepackage{mathrsfs}  

\def\eqref#1{equation~\ref{#1}}

\def\1{\bm{1}}

\DeclareMathAlphabet{\mathsfit}{\encodingdefault}{\sfdefault}{m}{sl}
\SetMathAlphabet{\mathsfit}{bold}{\encodingdefault}{\sfdefault}{bx}{n}

\title{Designing Biological Sequences via Meta-Reinforcement Learning and Bayesian Optimization}

\author{%
  Leo Feng\thanks{Correspondence to: Leo Feng <\href{mailto:leo.feng@mila.quebec}{leo.feng@mila.quebec}>
  \newline\hspace*{1.35em}\textsuperscript{1}CIFAR Senior Fellow,\ \textsuperscript{2}CIFAR AI Chair
  }
  \And
  Padideh Nouri
  \And 
  Aneri Muni
  \And
  Yoshua Bengio\,\textsuperscript{1}
  \And
  Pierre-Luc Bacon\,\textsuperscript{2}\And
  \newline
  \normalfont{Mila -- Universit\'{e} de Montr\'{e}al}
}

\begin{document}

\maketitle

\begin{abstract}
The ability to accelerate the design of biological sequences can have a substantial impact on the progress of the medical field. The problem can be framed as a global optimization problem where the objective is an expensive black-box function such that we can query large batches restricted with a limitation of a low number of rounds. Bayesian Optimization is a principled method for tackling this problem. However, the astronomically large state space of biological sequences renders brute-force iterating over all possible sequences infeasible. In this paper, we propose MetaRLBO where we train an autoregressive generative model via Meta-Reinforcement Learning to propose promising sequences for selection via Bayesian Optimization. We pose this problem as that of finding an optimal policy over a distribution of MDPs induced by sampling subsets of the data acquired in the previous rounds. Our in-silico experiments show that meta-learning over such ensembles provides robustness against reward misspecification and achieves competitive results compared to existing strong baselines.
\end{abstract}

\section{Introduction}
\label{submission}

Over the past few years, there has been a rapid interest in expediting the drug discovery process using Artificial Intelligence(AI). Leading pharmaceutical companies such as Pfizer, Sanofi and Roche are benefiting from AI in their pipelines to be able to keep up with the growing industry \cite{Mak2019}. In particular, designing biological sequences has been a long-standing challenge due to the large chemical space with very sparse functionally interesting sequences \cite{Capecchi2021}. 
The design of such sequences can be framed as a black-box optimization problem.
In this case, the time and cost-intensive wet-lab evaluation correspond to the objective black-box function.

Bayesian Optimisation is a principled way to tackle black-box optimization problems that approximates the expensive objective function (\textit{true oracle}) with a cheap surrogate model.
BO evaluates samples using an acquisition function to explore (i.e. select uncertain but informative sequences) or exploit (i.e. select sequences with high predicted value).
Although acquisition functions are relatively cheap to evaluate, naively searching over the space of possible sample locations can quickly become intractable.
In previous work, \citet{pmlr-v124-swersky20a} proposed to train a policy such that it modifies a population of sequences to directly optimise the acquisition function. 
\citet{belanger2019biological} propose to brute-force enumerate over the search space for shorter sequences and to use regularized evolution \cite{real2019regularized} for designing longer sequences. 
\citet{RomeroE193} propose to construct sequences with enhanced thermostability using crossover on an existing dataset of sequences and selects sequences using GP-UCB \cite{srinivas2009gaussian}.

In this work, we treat this discovery of biological sequence problem as training a generator to propose promising candidate sequences and use Bayesian Optimization as a selection procedure.
Generative models give us the ability to jump to the regions of the landscape that was never searched before.
In this paper, we propose a method called MetaRLBO where we frame the problem of training a generator as a Meta-Reinforcement Learning problem and apply Bayesian Optimization for batch black-box function optimization. 
In each round, we design our tasks by constructing a distribution of \textit{proxy oracles} and train our generator via Meta-Reinforcement Learning. 
Afterwards, we generate sequences with this meta-learned generator and finally select the sequences to query the \textit{true oracle} via Bayesian Optimization.
Our experiments show that MetaRLBO achieves competitive performance compared to existing baselines. 
In addition, we analyse the uncertainty estimates given by various surrogate models for batch sequence design.

    
    
    


\section{Background}

\subsection{Problem Setting}

The optimisation problem consist of generating a sequence that maximises the expected reward $s^*_t = \mathrm{argmax}_{s_t \in S} f(s_t)$. The sequence design can be viewed as a Markov decision process (MDP)  where $s_T$ is a sequence of length $T$ and of alphabet 
$\Sigma_{t=1}^T a_t$, that maximises an experimentally measured value (\textit{oracle}) $f: S \rightarrow \mathbb{R}$.
For example in the AMP task, $f(s)$ is defined as the antimicrobial efficacy of the candidate sequence s which can be computed using wet-lab experiments.
In particular, we are interested in a 
specific case of black-box optimisation: one which requires few rounds of evaluations over large batches, typical of wetlab experiments over biological assays (DNA, proteins, RNA, etc). This settings is one where in each round $r$ 
we want to minimize an objective function using the information of previously generated sequences. In each round, our algorithm generates a set of candidate sequences 
$\mathcal{B}_r = \bigcup_{i=1}^{B} s_i$
where $\mathrm{B}$ is the size of the batch. The sequences are then evaluated according to our oracle $f$ and added to the queried data $\mathcal{D} \leftarrow \mathcal{D} \bigcup \{(s, f(s)) | s \in \mathcal{B}_r \}$ and a new round is started. The objective is to maximise the score of aggregated sequences $\mathrm{max}_{s \in \bigcup_{i} \mathcal{B}_i} f(s)$.







\subsection{Meta-Reinforcement Learning}
The specific flavor of Meta-Reinforcement Learning (Meta-RL) considered in this paper is one where we have a distribution of Markov Decision Problems (MDPs) $\Omega(\mathcal{M})$ which we aim to control optimally. Each such MDP $\mathcal{M}_i \in \Omega(\mathcal{M})$ is defined over discrete state and action space $\mathcal{S}$ and $\mathcal{A}$ respectively. In the general case, each MDP can also have its own transition probability function $P_i: S \times \mathcal{A} \times \mathcal{S} \rightarrow \mathbb{R}$ and reward function $R_i: \mathcal{S} \times \mathcal{A} \rightarrow \mathbb{R}$. In our context, the problem structure is such that the transition function is known and shared across MDPs but the reward function varies.

We consider a performance criterion based on the expected discounted return over a finite horizon of length $H$ in searching for an optimal randomized policy $\pi: \mathcal{S} \to \text{Dist}(\mathcal{A})$. We write the expected return $\mathcal{J}$ under $\pi$:
$$
    \mathcal{J}(\pi) = \mathbb{E}_{P_i, \pi}\left[\sum_{t=0}^{H-1} \gamma^t r(s_t, a_t, s_{t+1}) \right] \enspace ,
$$
 $\gamma$ is a discount factor. 
 While dynamic programming \citep{Puterman1994} offer solution methods to solve finite-horizon problems, they are restricted to the set of deterministic policies and are only applicable to small known MDPs. Here, stochastic policies are specifically sought for due to the exploration problem.

In adopting a learning perspective, our meta-RL problem becomes one where we want to learn a  policy from the data gathered across a collection of MDPs sampled during training such that it generalises well to new MDPs $\mathcal{M}' \in \Omega(\mathcal{M})$ given little data. By generalization, we mean that we expect the learned policy to attain good performance in the new MDP with few \textit{adaptation} samples. 
The process of training a policy $\pi$ over sampled MDPs from the distribution of MDPs is \textit{meta-training} and that of evaluating the generalisation of $\pi$ is \textit{meta-testing}.

\textbf{Model Agnostic Meta-Learning (MAML)} 
MAML \cite{finn2017model} tackles a specific instance of the meta-learning problem \citep{bengio1991learning, schmidhuber1987} with a gradient-based algorithm inspired by bi-level optimization \citep{Bard1998}. More specifically, the MAML formulation considers the problem of learning initialisation parameters $\theta_0$ such that the policy adapts (improves) quickly within a few gradient updates to a new task.
More precisely, the meta-RL problem considered in MAML can be concisely formulated as:
\begin{align*}
    \min_{\theta_0}\,\, & \mathbb{E}_{\mathcal{M}' \sim \Omega(\mathcal{M})}[\mathcal{L}(\theta_K^{\mathcal{M}}; \mathcal{M}')] \\
    \text{s.t. } & \theta_{k+1} = \theta_k - \alpha \nabla_{\theta} \mathcal{L}(\theta_k^{\mathcal{M}}; \mathcal{M}) \\
    & \theta_0^{\mathcal{M}} = \theta \quad\quad \{\forall \mathcal{M} \in \Omega(\mathcal{M})\} \\ 
    & \mathcal{L} (\theta_k; \mathcal{M})= \frac{1}{T} \sum_{t} G_t \log \pi_{\theta_k} (a_t | s_t)
\end{align*}
where $K$ refers to the number of inner loop updates, $T$ is the number of timesteps (i.e. the horizon), $\theta$ refers to the meta-parameters, $\{(s_t, a_t, r_t)\}_{t=0}^{T-1}$ refers to a trajectory generated by the policy, and $G_t$ is the return from timestep $t$.



\subsection{Bayesian Optimization}

Bayesian Optimization (BO) aims to maximise a black-box (potentially non-differentiable) objective function in a few evaluations. 
It does so by building a surrogate model (e.g., a GP \citep{Rasmussen2006} or ensemble) of the true (oracle) objective function such that it can be queried at a lower cost. Given the surrogate model, BO then proceeds to efficiently compute the posterior distribution over the function scores and suggest promising candidates according to an acquisition function to be evaluated by the true oracle. 
Examples of acquisition functions are the following: Upper Confidence Bound (UCB) \cite{srinivas2009gaussian}, Posterior Mean, Thompson Sampling \cite{thompson1933on}, Entropy Search \cite{henning2011entropy, hernandez2014predictive}, or Expected Improvement \cite{jones1998efficient, mockus1975on} 
Acquisition functions not only take into account the score of a candidate as predicted by the surrogate model but also its uncertainty.
As a result, BO methods aim to find a suitable trade-off between exploration (gathering informative data) and exploitation (maximization of the black-box score function).

While our algorithm is compatible with any acquisition function, we found that Upper Confidence Bound (UCB) and Posterior Mean heuristics perform well in practice while being simple to implement. The first strategy, UCB, is defined as $\text{AF}_{\text{UCB}} = \mu(s) + \beta \sigma(s)$ where $s$ is a sample, $\mu(s)$ and $\sigma(s)$ are the mean and standard deviation predicted by the surrogate model and $\beta$ is a hyperparameter controlling the exploration-exploitation trade-off and with larger values favouring sequences of higher uncertainty.
In contrast, the Posterior Mean approach selects only sequences with high predicted scores.
As a result, the method does not explicitly try to explore uncertain areas. Posterior Mean can therefore be found as a subcase of UCB for the value of $\beta = 0$.


While Gaussian Processes have been the de facto choice in BO application, their poor scalability in high-dimensions \citep{wang2016bayesian} and over large datasets rendered them incompatible with modern deep learning tools. In this work, we use instead an ensemble of convolutional neural networks as the surrogate model. In this case, 
$\mu(s)$ is given by the mean of the predictions from the ensemble and $\sigma(s)$ is similarly estimated using the standard deviation.




\section{Methodology}


\subsection{Problem Setting}

\textbf{Sequence Generation as a Markov Decision Problem}  
In order for the BO procedure to scale over large batches of sequences with the fewest query to the oracle, we need to learn a probabilistic model allowing us to sample promising sequences according to their estimated score and degree of diversity. The requirement to be able to sample better candidates according to some specified objective function renders off-the-shelf probabilistic modelling inapplicable. Furthermore, it does not suffice to merely capture the statistical properties of a given dataset. A generator needs to be able to extrapolate beyond it and steer the BO towards promising regions of the solution space. 

We approach this problem from a reinforcement learning perspective and show that problem of learning a good generator can be seen as one of finding an optimal policy in an MDP with a specific structure. 
In this framework, our generator iteratively constructs new sequences by appending one element at a time to a string (initially empty) until it chooses to stop. 
We view this procedure as a sequential decision making problem where the action space is discrete (for the problems presented in this paper) and consists of all the possible symbols of a given alphabet (eg. amino acids) and the state space $\mathcal{S} = \bigcup_{t=1, \ldots L} \mathcal{A}^t$ is the set of all possible sequence prefixes where $L$ is the length of the sequence.
In other words, at time step $t$, the state is $s_t = a_0, \ldots a_{t-1} \text{ s.t. } a_i \in \mathcal{A}$. Because the state changes according to the action of the concatenation operation, the transition probability function in this MDP is fully known and deterministic, ie: $P(s_{t+1}|s_t, a_t) \triangleq \mathds{1}\left(s_{t+1} = \text{concat}(s_t, a_t)\right)$
In the current instantiation of our algorithm, we provide our system with feedback on the generated string only at the end of the sequence construction and not during. Note that if additional evaluative feedback is available for partial sequences, it could also be encoded within the reward function without any change to our algorithm. In all applications considered in this paper, we construct sparse reward functions which incorporate domain-specific scoring metrics as well as a general diversity bonus to encourage the generation of different enough sequences across rounds. 


\begin{figure*}
    \centering
    \includegraphics[width=0.7\textwidth]{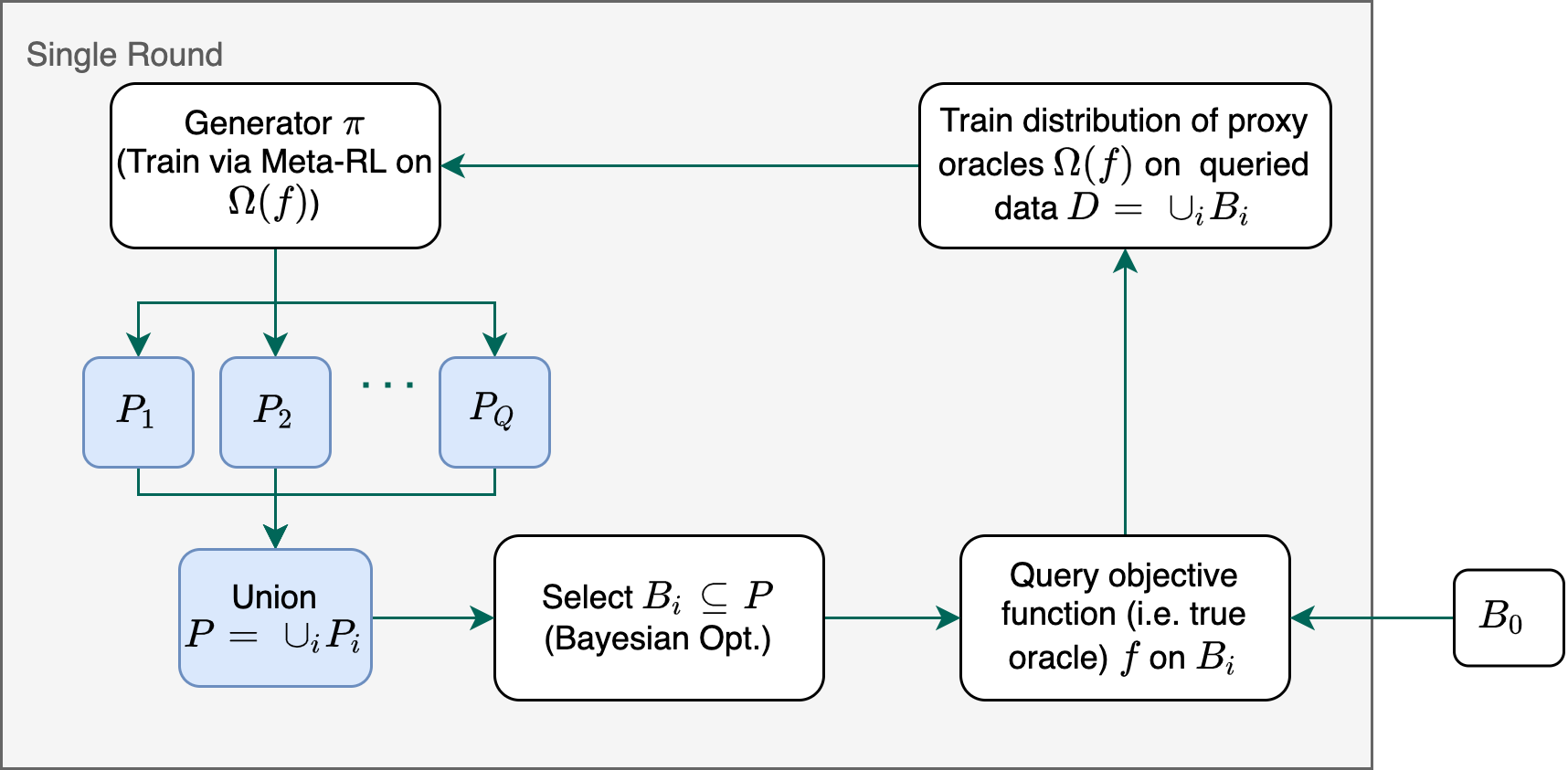}
    \caption{Schematic of the MetaRLBO Framework. 
    }
    \label{fig:schematic}
\end{figure*}

\subsection{Inner-Outer Loop Optimisation}
The nature of our problem prevents us from querying the oracle (wetlab) directly to evaluate the reward function. Hence, we devise a hierarchical approach in which a Bayesian optimisation procedure orchestrates the interaction with the oracle while using samples from the RL generator in the \textit{inner loop}. Figure \ref{fig:schematic} provides an overview of the interaction protocol between all components of our system. In this inner loop, the generator is trained over an ensemble of estimated reward functions acting as a proxy to the expensive true oracle. We are therefore under a model-based RL setting, but one where the transition dynamics are fully known and deterministic.

Learning an ensemble of reward functions offers many advantages: it provides uncertainty estimates readily usable by the BO procedure and provides more robustness in the presence of error in the estimated reward model. However, in order for this approach to be practical, we need to be able to learn good policies sufficiently fast for each such proxy oracle (reward function). We address this issue using a MAML-like meta-learning approach where we meta-learn initial parameters for the generator.


\subsubsection{Inducing a Distribution over Proxy MDPs}
We learn a distribution over proxy oracles $\Omega(f)$ by training on the queried dataset $D = \bigcup_i B_i$. The initial dataset $B_0$ can either be derived from previous experiments (e.g., related wet-lab experiments) or sampled randomly according to some specified prior distribution. 
In order to induce a distribution over possible MDPs, we sample different subsets of the current dataset  $D$, fit a proxy-oracle model to predict the score, and train an RL agent (generator) for each sub-sampled dataset of size $p|D|$ where $0 < p \leq 1$. 




\subsubsection{Training the generator via Meta-RL}
The particular flavor of meta-learning considered in this paper is based on the adaptation procedure of \cite{finn2017model}. In each meta-training step, we sample a set of $V$ proxy oracles $\hat{f}_1, \hat{f}_2, \ldots \hat{f}_V \in \Omega(f)$. Each proxy oracle specifies a different MDP (i.e. task) $M_i$ for which the reward function is given by $\hat{f}_i$.
During the adaptation phase of our generator, we fine-tune $\pi_\theta$ for $K$ steps using a REINFORCE gradient estimator from a learned $\theta^{0}$. 
\begin{align*}
    \theta^i_{t+1} \leftarrow \theta^i_t -  \alpha \nabla_{\theta} \mathcal{L}(\theta^i_t ; M_i) \enspace ,
\end{align*}
In order to learn this initialisation, we backpropagate through the $K$ gradient steps as follows:
\begin{align*}
    \theta^{0}_{t+1} \leftarrow \theta^0_{t} -  \gamma \frac{1}{V} \sum_{i=1}^V \nabla_{\theta} \mathcal{L}(\theta^i_K ; M_i) 
\end{align*}
While higher-order differentiation of the REINFORCE surrogate loss results in a biased meta-gradient \citep{Foerster2018}, we choose to ignore this technicality in favour of increased computation efficiency as commonly done in practice with other algorithms. 

\subsubsection{Generating sequences}
When generating sequences to evaluate, we sample a number of proxy oracles $\bar{f}_1, \bar{f}_2, \ldots, \bar{f}_Q \in \Omega(f)$, each reflecting a belief of the true oracle $f$. We then finetune each generator for some pre-defined number of steps for each tasks, giving us a set of policies $\pi_{\theta^{(1)}}, \pi_{\theta^{(2)}}, \ldots \pi_{\theta^{(Q)}}$. 
 At this end of this procedure, we then query $|P|$ sequences from each policy 
 that we gather into a batch
$\mathcal{P} = \bigcup_{i=1, \ldots, Q} P_i$.
Since the reward function of each MDP is different, generated sequences from different policies $\pi_{\theta^{(1)}}, \pi_{\theta^{(2)}}, \ldots \pi_{\theta^{(Q)}}$ provide uncertainty about the generation process as a whole.


\textbf{Promoting Diversity within a round}
In each round of interaction at the outer level (from BO to wetlab), a batch of candidate sequences to be presented to the Bayesian Optimization must be formed using the ensemble of generators. By virtue of using a different reward function in each MDP, the corresponding learned generators tend to be different from each other, i.e. the learned optimal policies need not be the same. Hence, the combination of subset sampling and meta-learning readily provides a mechanism for generating more diverse candidates in each round. That is: it provides a form of meta-exploration on top of the built-in exploration mechanism of softmax (randomized) policies. 


\textbf{Promoting Diversity Across Rounds} 
Similar to the existing literature on count-based methods and density models for promoting exploration \cite{Bellemare2016,  Angermueller2020Modelbased}, we augment the reward with an exploration bonus. This method encourages the generation of diverse sequences relative to the ones queried in the previous rounds (\textit{diversity across rounds}).
This augmented reward is defined as $r = f(s) - \lambda \mathrm{dens}_{\epsilon}(s)$ where $\mathrm{dens}_{\epsilon}(s)$ is the weighted number of sequences with a distance less than a threshold $\epsilon$ from $s$. 
$\lambda$ is a hyperparameter that controls the the strength of exploration.

\subsubsection{Selecting via Bayesian Optimization}


When selecting sequences, we evaluate the generated sequences $P$ according to an acquisition function. 
For the surrogate model, we can re-use the proxy models ($\bar{f}_1, \ldots \bar{f}_Q$) sampled from the proxy oracle distribution $\Omega(f)$ that were used for training our policies. 
Aggregating the proxy models together, we get a surrogate model comprising of an ensemble of neural networks without the need for additional computation.
Alternatively, we can train a new surrogate model given the queried dataset $\bigcup_i B_i$.
However, in our experiments, we found that re-using the proxy models works well in practice and provides good uncertainty estimates.
For evaluation, we greedily select the sequences $B_i \subseteq P$ that maximise the acquisition function and query them using the objective function (i.e. true oracle) $f$.




\section{Experiments}



\subsection{Training details}

\textbf{Policy Network}
We implement the policy as a feedforward neural network that takes as input the flattened one-hot encoding of the generated string, i.e. a vector of dimension $\mathbf{R}^{L \times A}$ where $L$ is the length of the sequence and $A$ is the size of the alphabet. The network output is the logits for the distribution over the next character to be generated.
We also use entropy bonus to avoid premature convergence \citep{ahmed2019understanding}. 
We apply positional encoding \cite{vaswani2017attention} to the one-hot representation of the input string.

\textbf{Surrogate Model} 
In our experiments, we use an ensemble of CNNs for our proxy models. 
For each ensemble member, the model is trained for 10 epochs using the ADAM optimizer \cite{kingma2015adam} with a batch size of 50 and mean squared error loss.
In practice, we found setting $p=1.0$ sufficient for training neural networks since diversity can be induced by the different neural network initialisation.
However, setting $p < 1.0$ is necessary for evaluating different proxy models (see Section \ref{subsection:analysis})




\textbf{Training}
During meta-training, we sample $V=4$ proxy models per meta-updates.
When generating sequences, we sample $Q=32$ proxy oracles and generate $|P|=64$ sequences from each of $\pi_\theta^{(i)}$. 
In our experiments, we assume no prior knowledge is given, generating $B_0$ via a random policy.

\subsection{Datasets}

We test our approach on three kinds of sequence optimization problems: designing antimicrobial peptide sequences (AMPs), ribonucleic acid (RNA) sequences, and sequences that maximise a synthetic Alternating Ising Model. 
In the AMP problem, the sequences are of variable lengths.
In the Alternating Ising and the RNA14 Task, the sequence length is fixed.

\textbf{Antimicrobial Peptides (AMP)}
AMPs are small peptides with amino acids as their building blocks and a length generally ranging from 8 to 75 (amino acids). 
This task consists of an alphabet of size 20 and a maximum length of 50, which gives rise to a search space of size $O(20^{50})$.
Following \citep{Angermueller2020Modelbased}, we train a random forest classifier to predict whether a sequence is antimicrobial towards a certain pathogen and we use that as our ground truth (wet-lab) simulator.
In our experiments, we perform 12 rounds with a batch size of 250.

\textbf{Alternating Ising Model}
We consider the synthetic problem of generating a string of alternating characters \citep{pmlr-v124-swersky20a}. The string with the highest score is one that only alternates between two characters. We consider the problem setting with lengths of 20
with an alphabet of size 20. 
As such, the search space of this problem is $O(20^{20})$.
In our experiments, we perform 16 rounds with a batch size of 500.

\textbf{Ribonucleic Acid (RNA)}
We design RNA sequences with length 14 from the alphabet of size 4 nucleotides, this would give us a search space of $O(4^{14})$. Here the optimisation problem can be defined as finding a sequence that maximises the negative binding energy towards a hidden RNA target of length 50. To simulate the ground truth oracle we use FLEXS package. \cite{sinai2020adalead, lorenz2011viennarna}
In our experiments, we perform 12 rounds with a batch size of 100.



\subsection{Baselines}

In our experiments, we compare against several baselines including exploration algorithms proposed in FLEXS \cite{sinai2020adalead} and  Bayesian Optimization methods proposed in \cite{pmlr-v124-swersky20a}. Primarily, we consider: (1) \textbf{Random}: A baseline that mutates a random previously measured sequence. (2) \textbf{Genetic}: A naive genetic algorithm that uses a Wright-Fisher model and single point mutations and recombinations. \cite{sinai2020adalead}. (3) \textbf{CMA-ES}: Covariance Matrix Adaptation Evolution Strategies \cite{hansen2016cma, sinai2020adalead}. (4) \textbf{DynaPPO}: A model-based RL algorithm that learns a reward function given by the mean predicted value of an ensemble of various different models such as random forests, gaussian processes, and bayesian ridge regression. \cite{Angermueller2020Modelbased, sinai2020adalead}. (5) \textbf{Adalead}: A model-guided evolutionary greedy algorithm \cite{sinai2020adalead}.


\subsection{Comparison with baselines}

\begin{figure*}[t]
    \centering
    \begin{subfigure}[t]{0.32\textwidth}
        \centering
        \includegraphics[width=\textwidth]{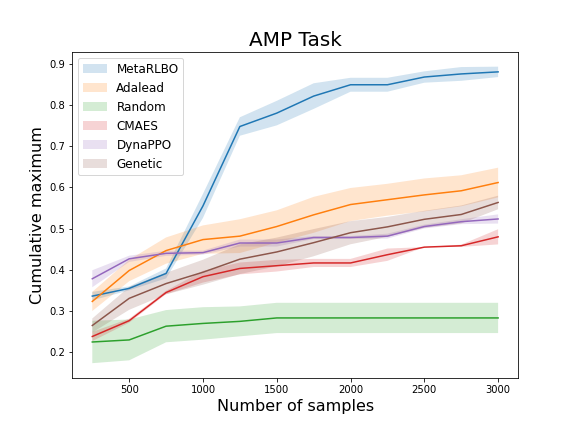}
        \caption{AMP}
        \label{fig:AMP_multiround}
    \end{subfigure}
    \hfill
    \begin{subfigure}[t]{0.32\textwidth}
        \centering
        \includegraphics[width=\textwidth]{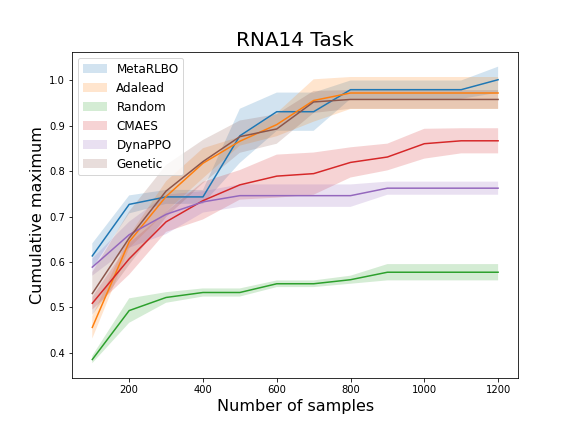}
        \caption{RNA14}
        \label{fig:RNA14_multiround}
    \end{subfigure}
    \hfill
    \begin{subfigure}[t]{0.32\textwidth}
        \centering
        \includegraphics[width=\textwidth]{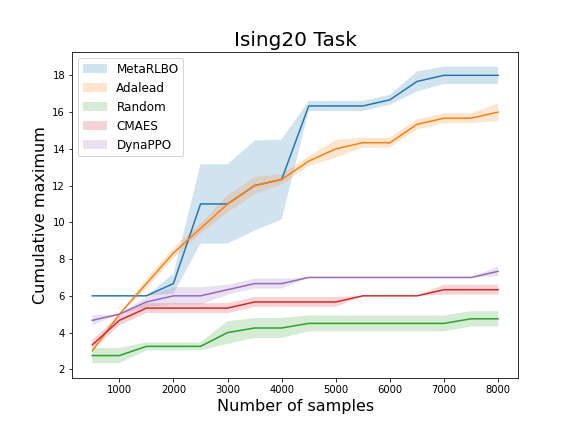}
        \caption{Ising20}
        \label{fig:Ising20_multiround}
    \end{subfigure}
    \caption{Cumulative maximum scores given by the objective function $f$. (a) Antimicrobial Peptide Design (b) Ribonucleic Acid (c) Alternating Ising Model}
    \label{fig:multiround_results}
\end{figure*}

For a fair comparison, mutative methods are initialised with a random string, ensuring no prior knowledge is given to the method.
Figure \ref{fig:AMP_multiround} shows results using MetaRLBO on the AMP, RNA, and Alternating Ising Model problems. 
In Table \ref{table:ising_method_comparison}, we compare our method with Bayesian Optimization methods and evolutonary methods.
We see that MetaRLBO outperforms several existing baselines. 





\begin{table}[t]
    \begin{minipage}{.5\linewidth}
        \begin{tabular}{l|ll}
        \hline
        Alternating Chain & \multicolumn{2}{c}{Length 20}  \\ \hline
        Acq. Function & UCB & POST \\ \hline
        Single Mutant & 14.67 & 14.67\\
        Regularized Evol & 14.67 & 14.67\\
        BO + Single Mutant & 15.33 &13.67\\
        BO + Regularized Evol & 16.67 &15.00\\
        BO + DES & 16.67 & 16.33\\
        \hline
        \textbf{MetaRLBO} & \textbf{18.00} & \textbf{17.00}
        \end{tabular}
        \centering
        \caption{Alternating Chain Ising20 task. Previous results obtained from \cite{pmlr-v124-swersky20a}}
        \label{table:ising_method_comparison}
    \end{minipage}
    \begin{minipage}{.5\linewidth}
        \begin{tabular}{c|c}
        \hline
        Proxy Model & Cumul. Max \\ \hline
        CNN & 18 \\ \hline
        MLP & 15 \\ \hline
        \end{tabular}
        \caption{Ising20 Task using UCB. We compare the performance of MetaRLBO when trained on different proxy oracles.}
        \label{table:ising20_proxy_model}
    \end{minipage}
\end{table}

\subsection{Analysis}
\label{subsection:analysis}


\begin{figure*}[t]
    \centering
    \begin{subfigure}[t]{0.32\textwidth}
        \centering
        \includegraphics[width=\textwidth]{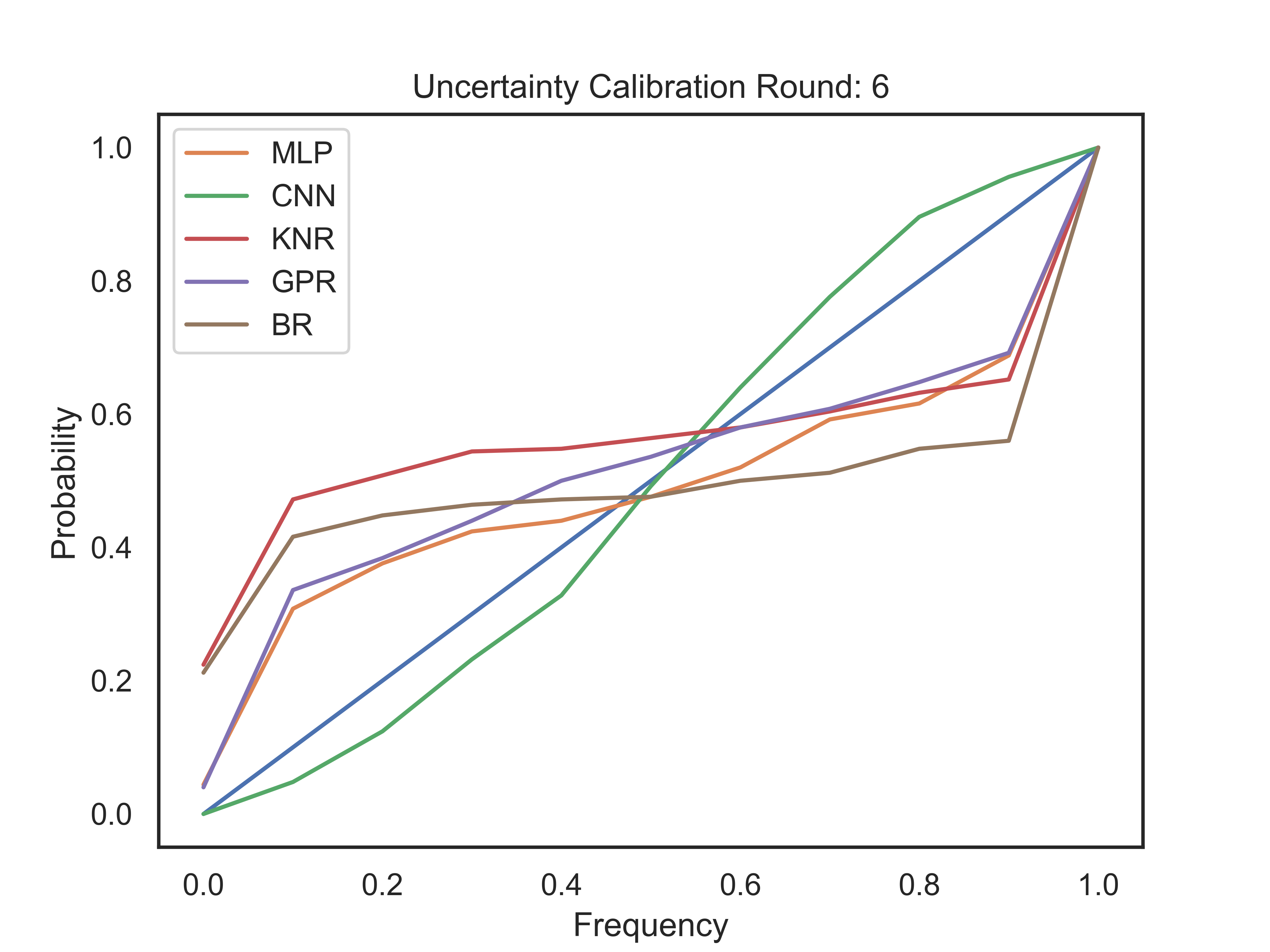}
        \caption{Uncertainty Calibration Plot}
        \label{fig:uncertainty_calibration}
    \end{subfigure}
    \hfill
    \begin{subfigure}[t]{0.32\textwidth}
        \centering
        \includegraphics[width=\textwidth]{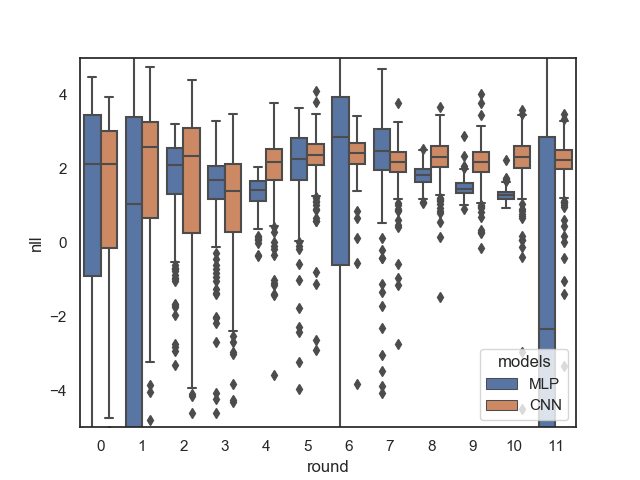}
        \caption{Box Plot of NLL}
        \label{fig:amp_nll}
    \end{subfigure}
    \hfill
    \begin{subfigure}[t]{0.32\textwidth}
        \centering
        \includegraphics[width=\textwidth]{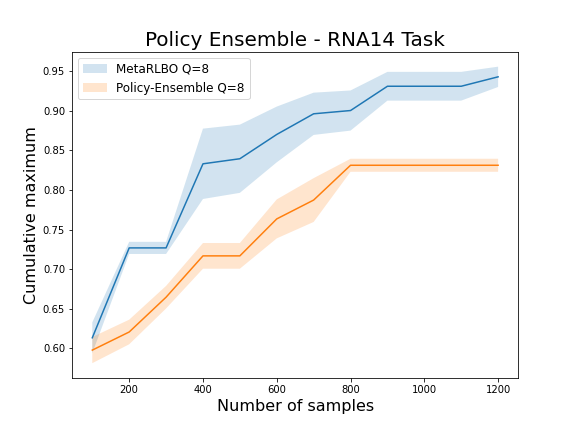}
        \caption{Ensemble of Policies}
        \label{fig:ablation_policy_ensemble}
    \end{subfigure}
    \caption{(a) Uncertainty Calibration Plot comparing the expected vs. empirical rates of observing an outcome $y_t$.
    A perfect model would achieve the line $y=x$. 
    (b) Box Plot of the Negative Log Likelihood comparing an ensemble of MLP and an ensemble of CNNs. The surrogate models are trained on data selected in the first $n$ rounds and evaluated on sequences selected in the $n+1$-th round. (c) An ablation comparing an ensemble of fine-tuned policies from a meta-learned initialization and an ensemble of policies trained from scratch.}
    \label{fig:uncertainty_analysis}
\end{figure*}



\textbf{Uncertainty Model} 
Prior work considered using an ensemble of MLPs as the surrogate model for Bayesian Optmization \cite{belanger2019biological, pmlr-v124-swersky20a}
In Figure \ref{fig:uncertainty_calibration}, we show an uncertainty calibration plot comparing various surrogate models trained on sequences selected in the first $n$ rounds and evaluated on the sequences selected in the $n+1$-th round. 
This is a measure of how well a surrogate model is at estimating uncertainty in a multi-round setting.
Calibration in the regression setting \cite{kuleshov2018accurate} means that $y_t$ should fall in a $c\%$ confidence interval $c\%$ of the time.
This means that in Figure \ref{fig:uncertainty_calibration}, we would want surrogate models to achieve as close to $y=x$ as possible.

To generate the data for the analysis, we run MetaRLBO on the AMP task using UCB as our acquisition function and an ensemble of CNNs as our surrogate model.
In Figure \ref{fig:uncertainty_calibration}, we considered surrogate models such as an ensemble of feedforward neural networks with three layers (32, 8, 4 units and $p=1.0$) used in \citet{belanger2019biological} and \citet{pmlr-v124-swersky20a}, ensemble of convolutional neural networks ($p=1.0$), ensemble of Bayesian Ridge Regressors ($p=0.8$), ensemble of K-Neighbors Regressors ($p=0.8$), and Gaussian Processes. Additional plots are included in the Appendix.
The results show an an ensemble of CNNs is better calibrated than the other surrogate models. 
For a more comprehensive analysis, we also measure the Negative Log-likelihood, another popular metric for evaluating predictive uncertainty \cite{candela2006evaluating}. 
We see that an ensemble of MLP and an ensemble of CNN are within intervals of each other under this metric. However, the ensemble of MLP has higher variance.

We also evaluate the uncertainty models empirically in practice on the Alternating Ising Model task.
In this experiment, we run MetaRLBO from scratch using either MLP or CNN as the proxy model $\Omega(f)$.
In practice, we found CNNs to perform better (see Table \ref{table:ising20_proxy_model}).
The result of our analysis suggests that convolutional neural networks are better suited for uncertainty estimation in biological sequence design than that of feedforward neural networks.

\textbf{Ensemble of Policies}. 
As an ablation, we compare MetaRLBO with training an ensemble of policies.
Instead of fine-tuning different $Q$ policies from a set of meta-trained parameters, we train $Q$ different policies from scratch per round such that each policy optimises for a different proxy oracle.
In these experiments, we set $Q=8$, generating $256$ sequences from each policy.
In our MetaRLBO experiments, we additionally set $V=4$ and $K=2$.
Empirically, we found that training a policy via meta-reinforcement learning improved the performance (see Figure \ref{fig:ablation_policy_ensemble}).
\section{Related Work}
\label{04:rel_work}


Machine learning has been effective in optimizing DNA and protein sequences \cite{Wang563775, chhibbar2019generating, dejongh2019designing}.
We can divide existing methods for biological sequence design into three categories \cite{Angermueller2020Modelbased}
: directed evolution \cite{Yang2019a}, optimization using discriminative models (e.g. Bayesian Optimization), and optimization using generative model \cite{VanOort2021, Muller2018}.

Optimization using discriminative models alternates between 1) fitting a proxy model $f'(x)$ to approximate $f(x)$ and 2) selecting a batch of sequences using an acquisition function based on $f'(x)$.
\citet{pmlr-v124-swersky20a} introduces Deep Evol. Solver (DES) to train a policy to evolve a population of strings with character-level edits to optimize an acquisition function.
Evolutionary methods perform a local search in the space of sequences, in a hill-climbing fashion. As a result, they have poor sample efficiency.
Directed evolution which mimics the natural selection involves the generation of a large set of diverse candidates to find functionally interesting candidates \cite{Yang2019a}. 
Optimization methods based on generative models learn a generator $p_\theta(x)$ that aims to maximise the expected value of $f(x)$: $\mathbf{E}_{x \sim P_\theta(x)}[f(x)]$. 
One way to optimize this is via RL.
\citet{wang2021deep} propose to train a long short-term memory (LSTM) generative model and a bidirectional LSTM classification model for designing AMP sequences. 
\citet{schuchardt2019learning} propose to train an RL agent how to mutate, perform selection, and cross-over on sequences.
DynaPPO \cite{Angermueller2020Modelbased} is a model-based RL algorithm that trains an agent to generate sequences that maximise the predicted mean given by an ensemble of various models such as Random Forest, Bayesian Ridge Regression, MLP, etc. 
Our proposed method (MetaRLBO) lies in the intersection between optimization using discriminative models and optimization using generative models.
As far as we are aware, we are the first to propose combining Meta-RL and Bayesian Optimization for biological sequence design.




\section{Conclusions and Future Work}

We have introduced MetaRLBO, a method that combines a generator based on Meta-Reinforcement Learning and a selection procedure via Bayesian Optimization to propose promising sequences that optimise a black-box function. 
We also analyse the uncertainty estimates of surrogate models and find that an ensemble of CNNs is better calibrated than an ensemble of MLPs as used in prior work.



When selecting sequences, MetaRLBO currently selects sequences greedily according to the acquisition function.
However, instead of querying two promising but similar sequences, it would be advantageous to query sufficiently different sequences to explore the search space.
This additional level of exploration can be tackled with batch-aware bayesian optimization methods such as BatchBald \cite{kirsch2019batchbald} and would be an interesting future direction to explore.



\bibliography{main}

\newpage
\appendix

\section*{Appendix}

\begin{figure*}[ht]
    \centering
    \begin{subfigure}[t]{0.32\textwidth}
        \centering
        \includegraphics[width=\linewidth]{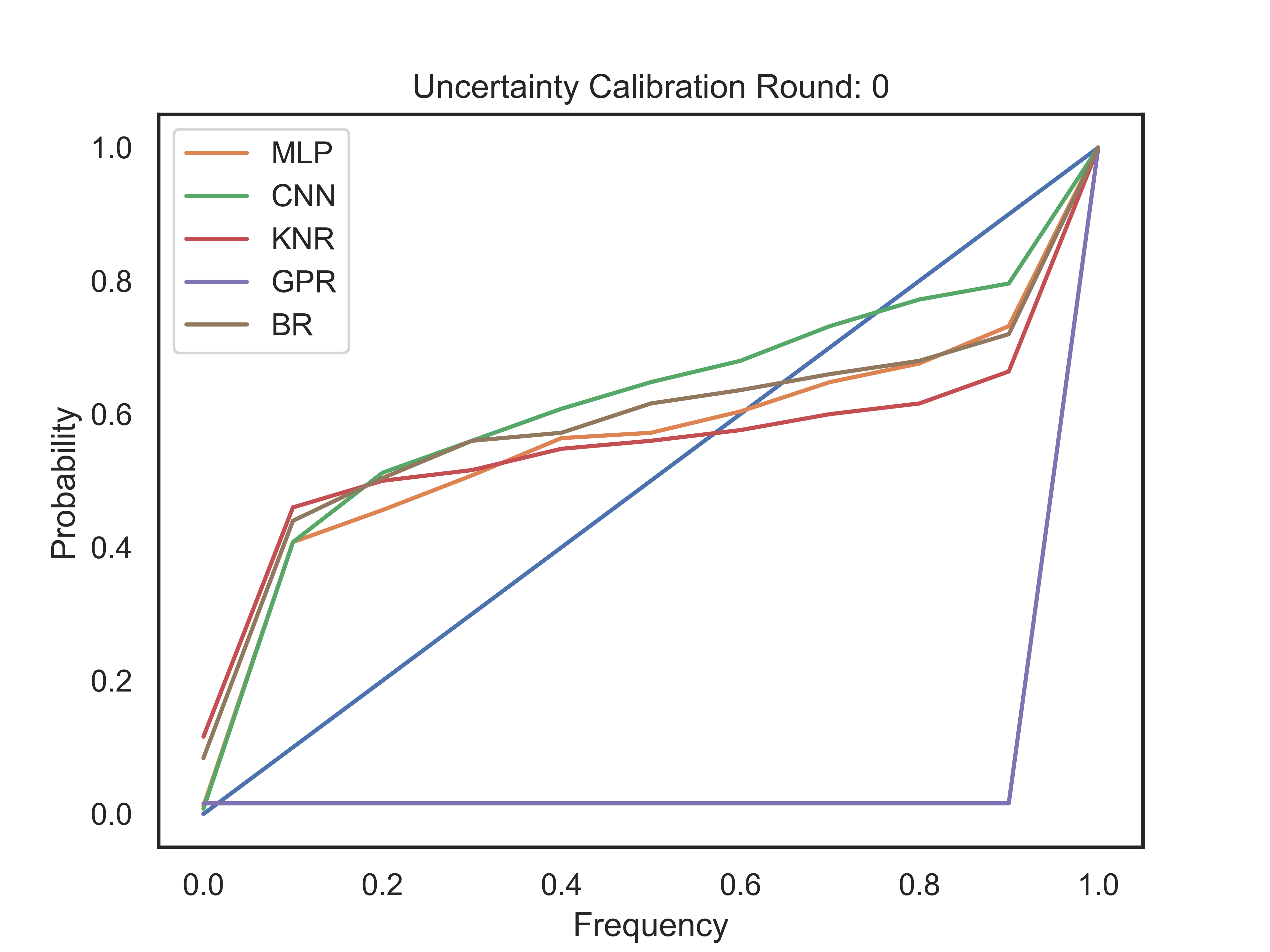}
    \end{subfigure}
    \hfill
    \begin{subfigure}[t]{0.32\textwidth}
        \centering
        \includegraphics[width=\linewidth]{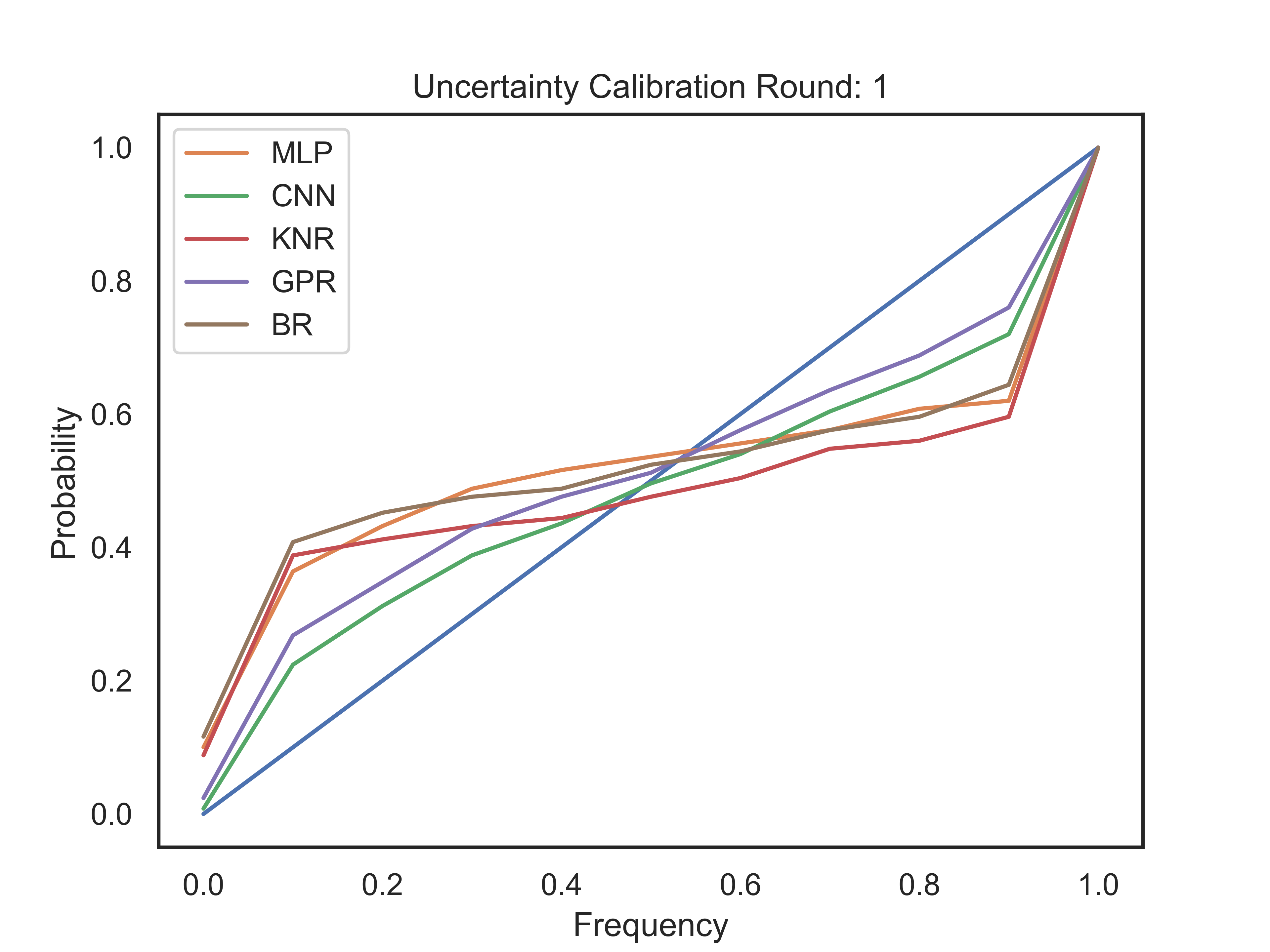}
    \end{subfigure}
    \hfill
    \begin{subfigure}[t]{0.32\textwidth}
        \centering
        \includegraphics[width=\linewidth]{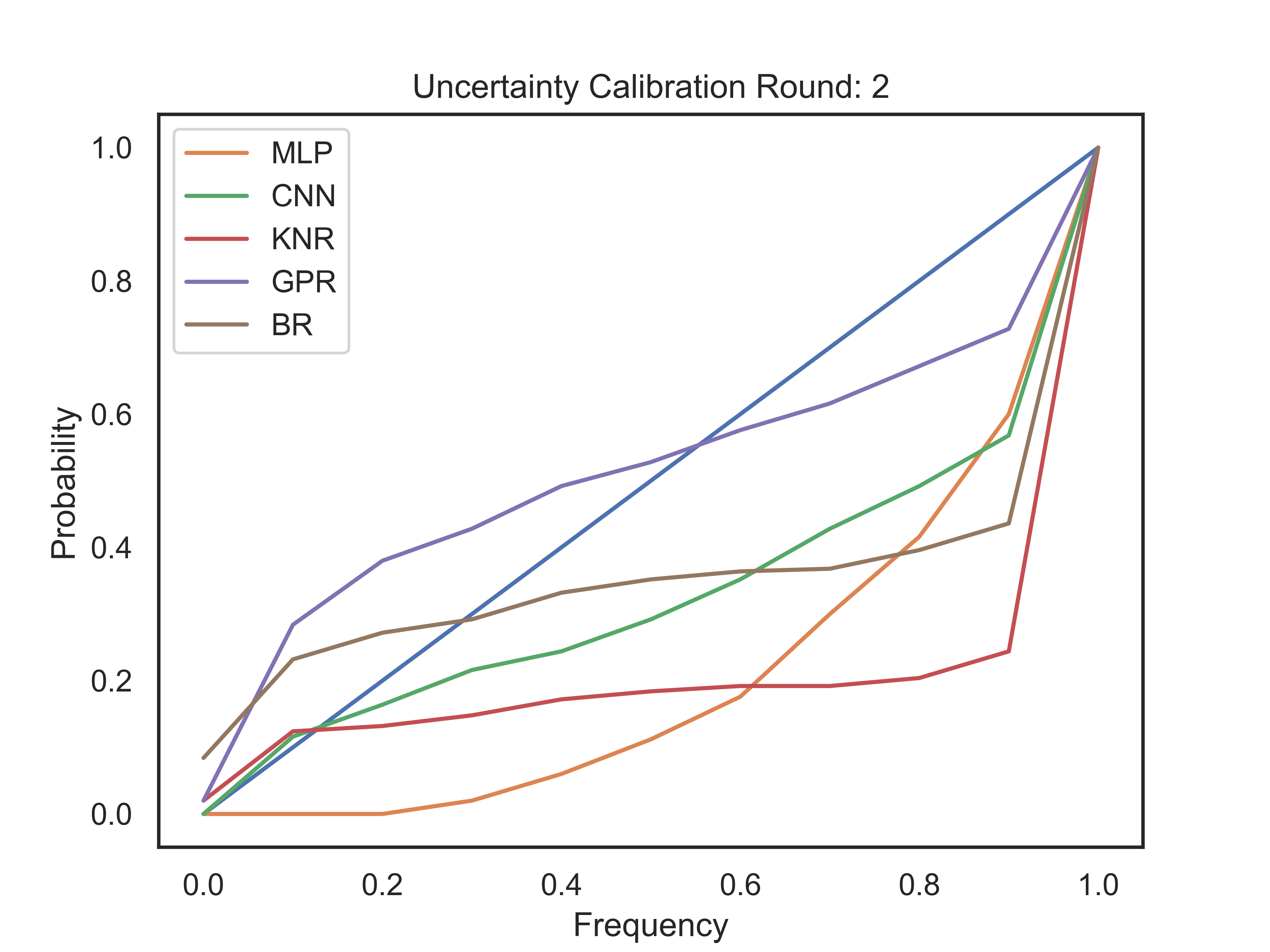}
    \end{subfigure}
    \hfill
    \begin{subfigure}[t]{0.32\textwidth}
        \centering
        \includegraphics[width=\linewidth]{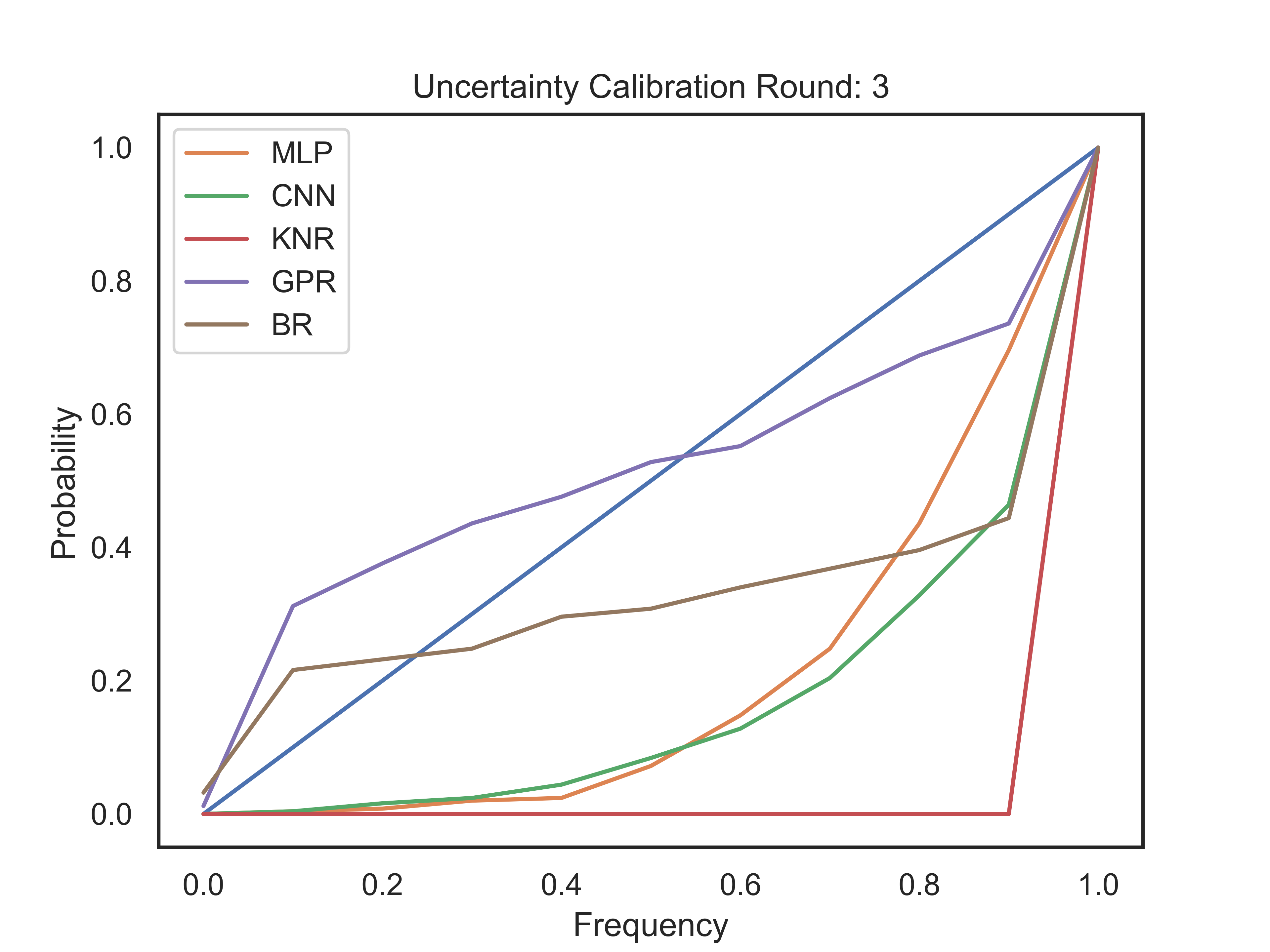}
    \end{subfigure}
    \hfill
    \begin{subfigure}[t]{0.32\textwidth}
        \centering
        \includegraphics[width=\linewidth]{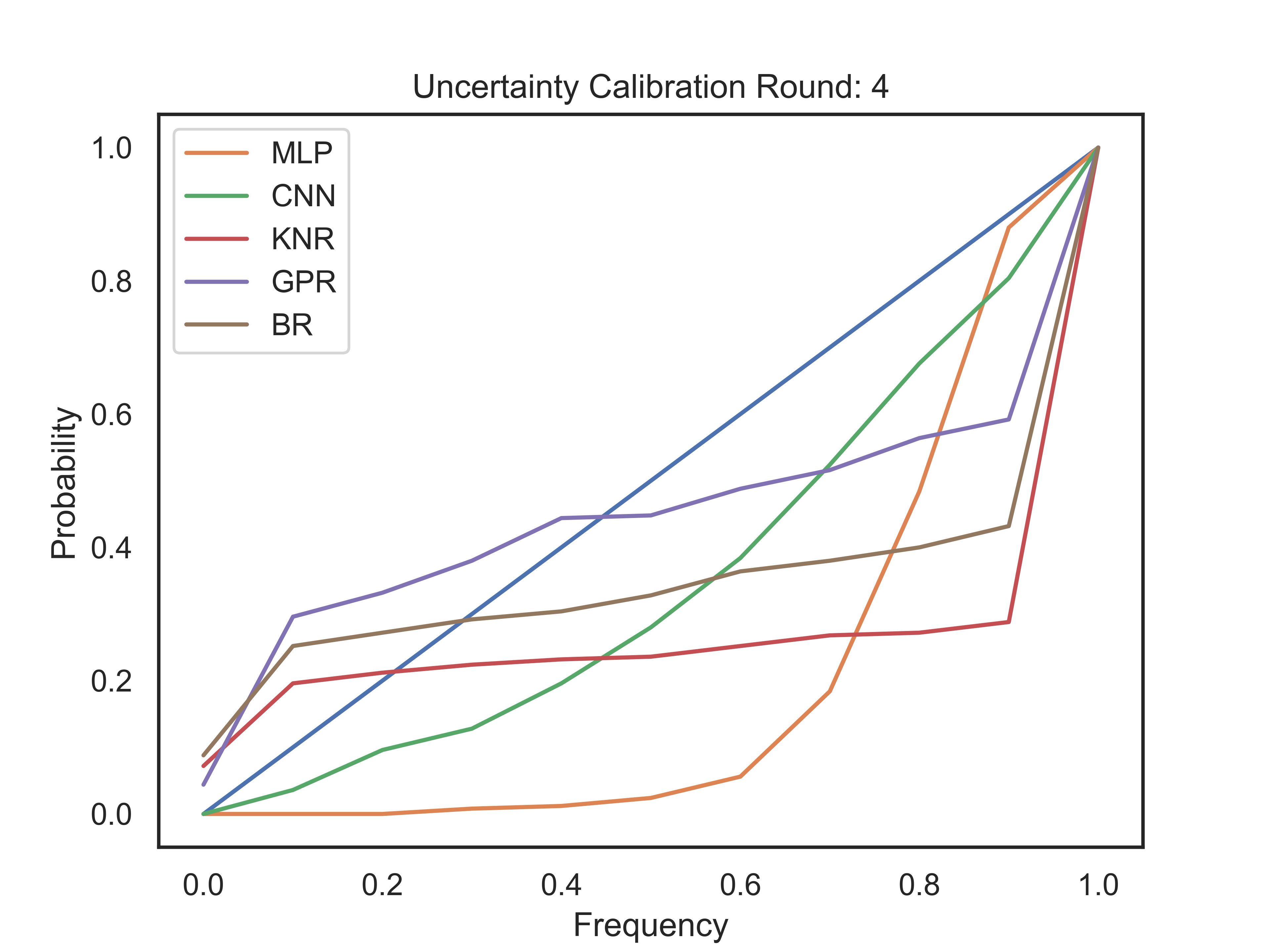}
    \end{subfigure}
    \hfill
    \begin{subfigure}[t]{0.32\textwidth}
        \centering
        \includegraphics[width=\linewidth]{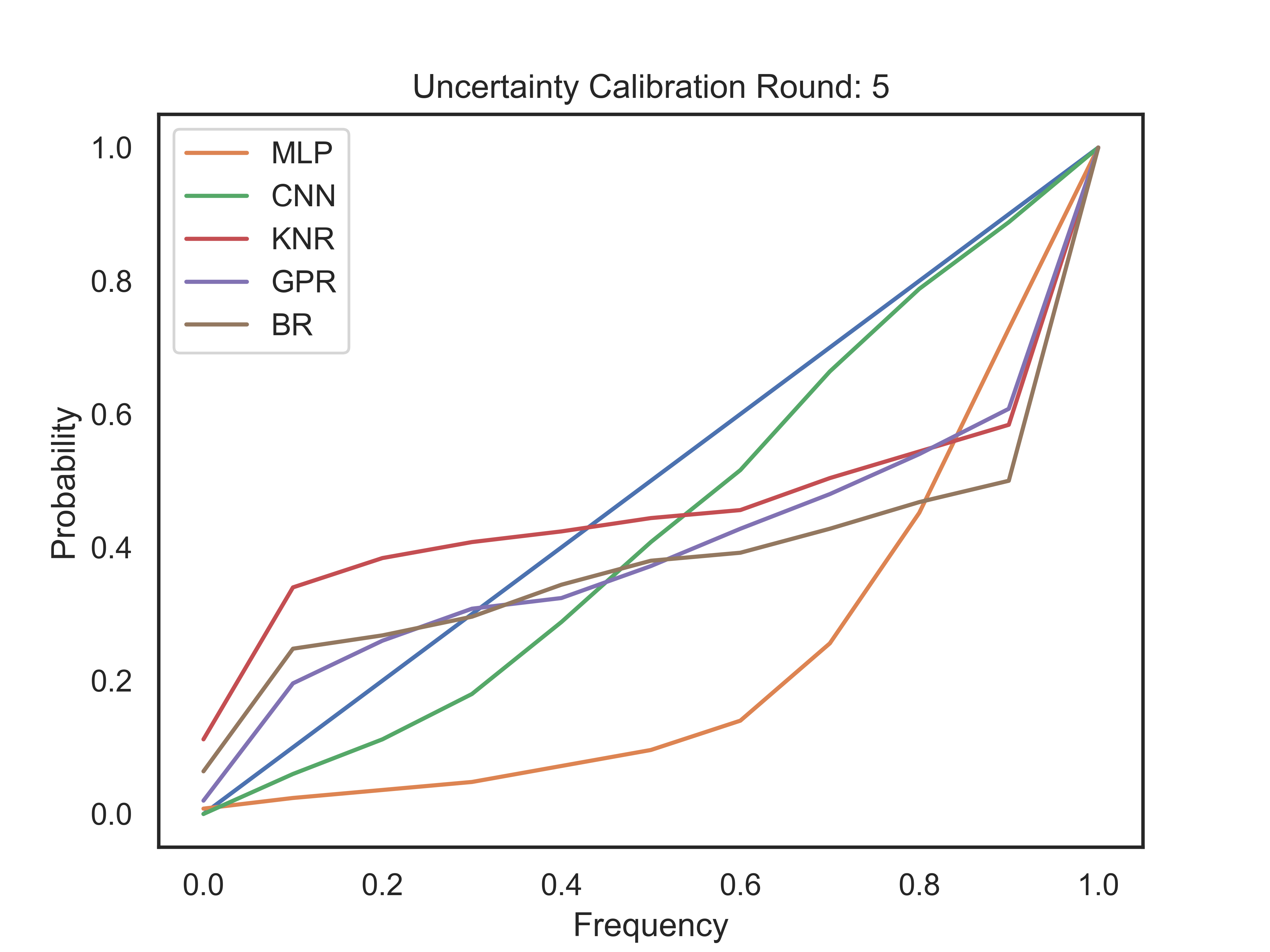}
    \end{subfigure}
    \hfill
    \begin{subfigure}[t]{0.32\textwidth}
        \centering
        \includegraphics[width=\linewidth]{imgs/uncertainty_calibration/Uncertainty_Calibration_Round-6.png}
    \end{subfigure}
    \hfill
    \begin{subfigure}[t]{0.32\textwidth}
        \centering
        \includegraphics[width=\linewidth]{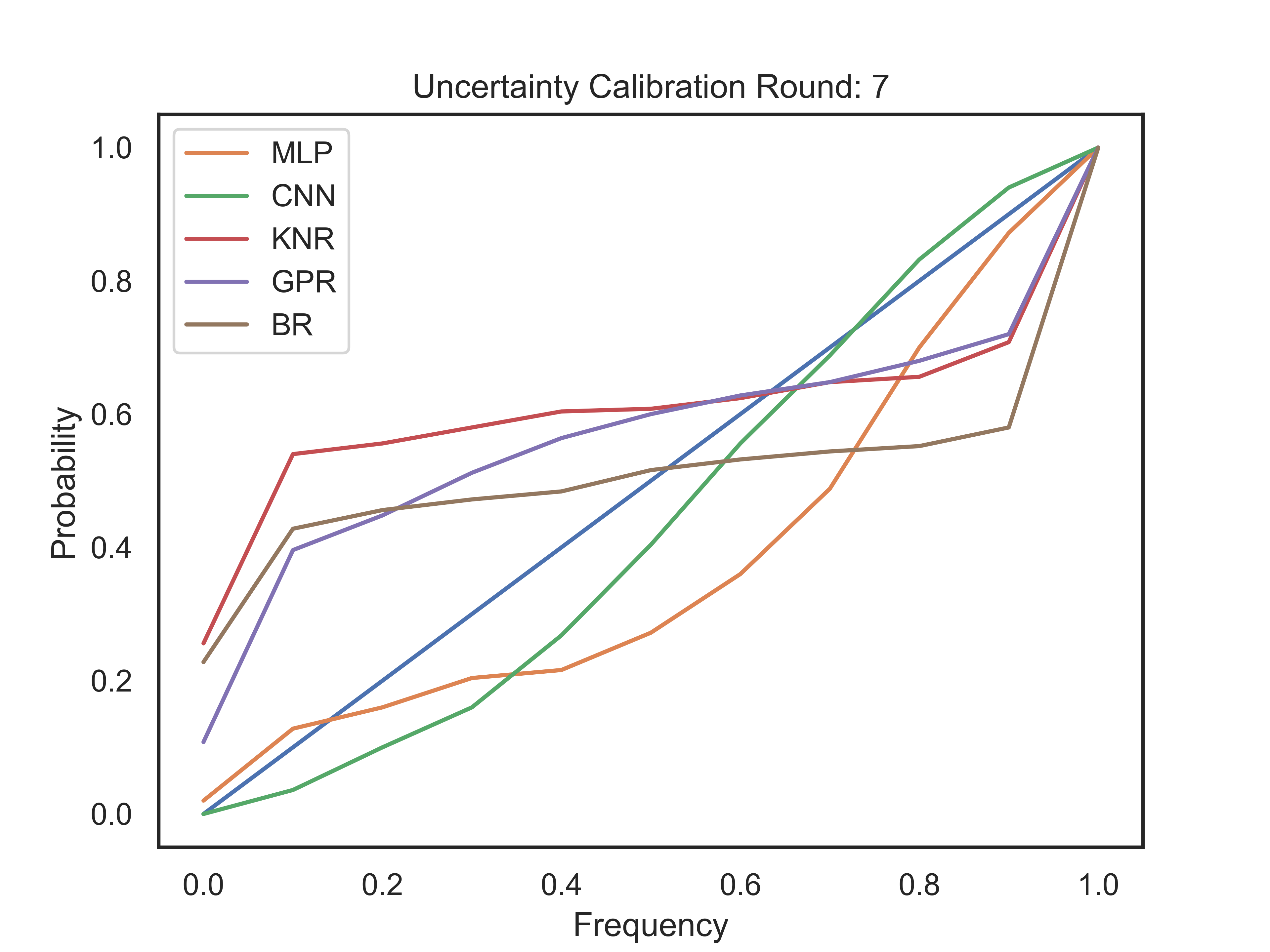}
    \end{subfigure}
    \hfill
    \begin{subfigure}[t]{0.32\textwidth}
        \centering
        \includegraphics[width=\linewidth]{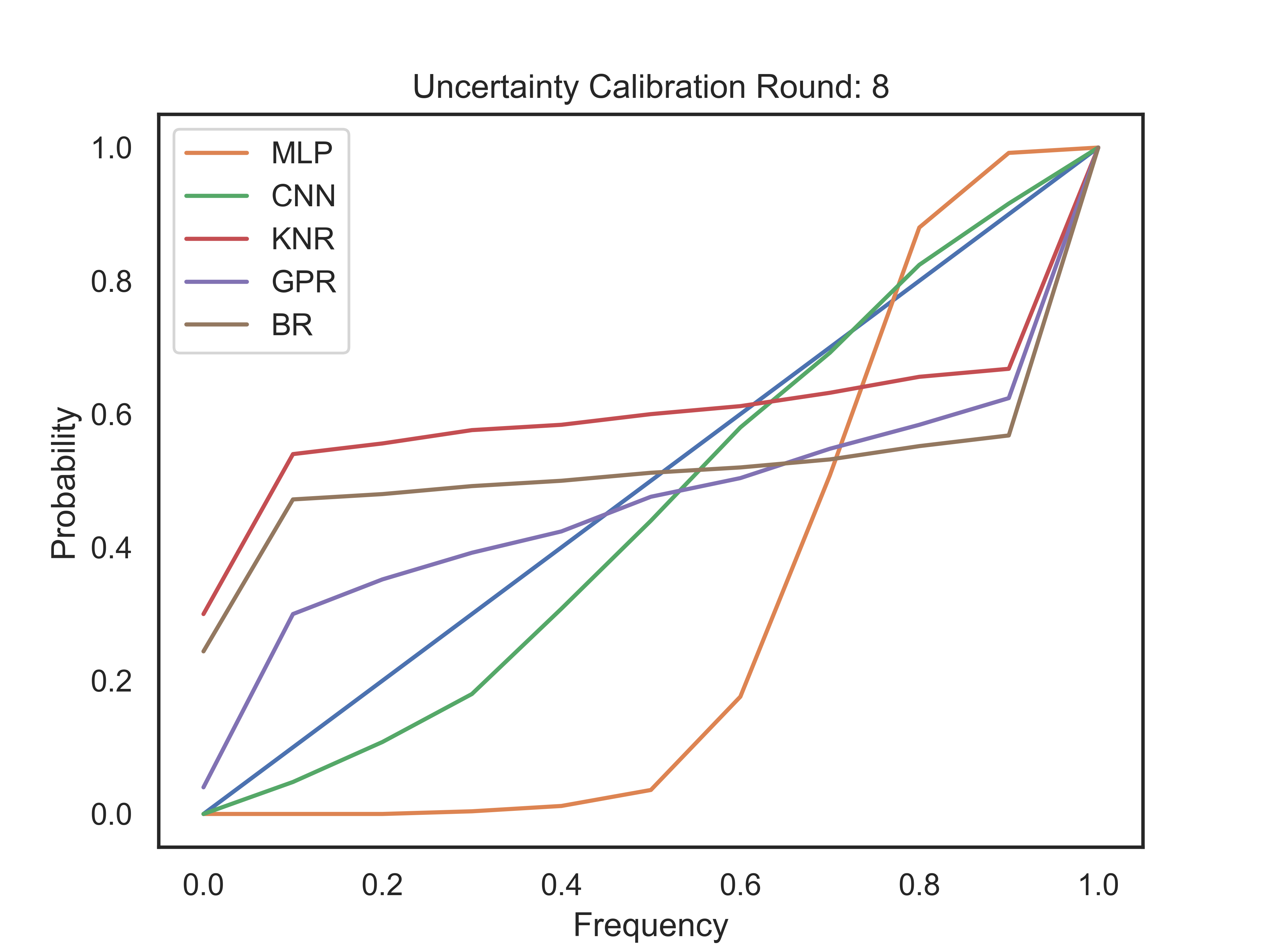}
    \end{subfigure}
    \hfill
    \begin{subfigure}[t]{0.32\textwidth}
        \centering
        \includegraphics[width=\linewidth]{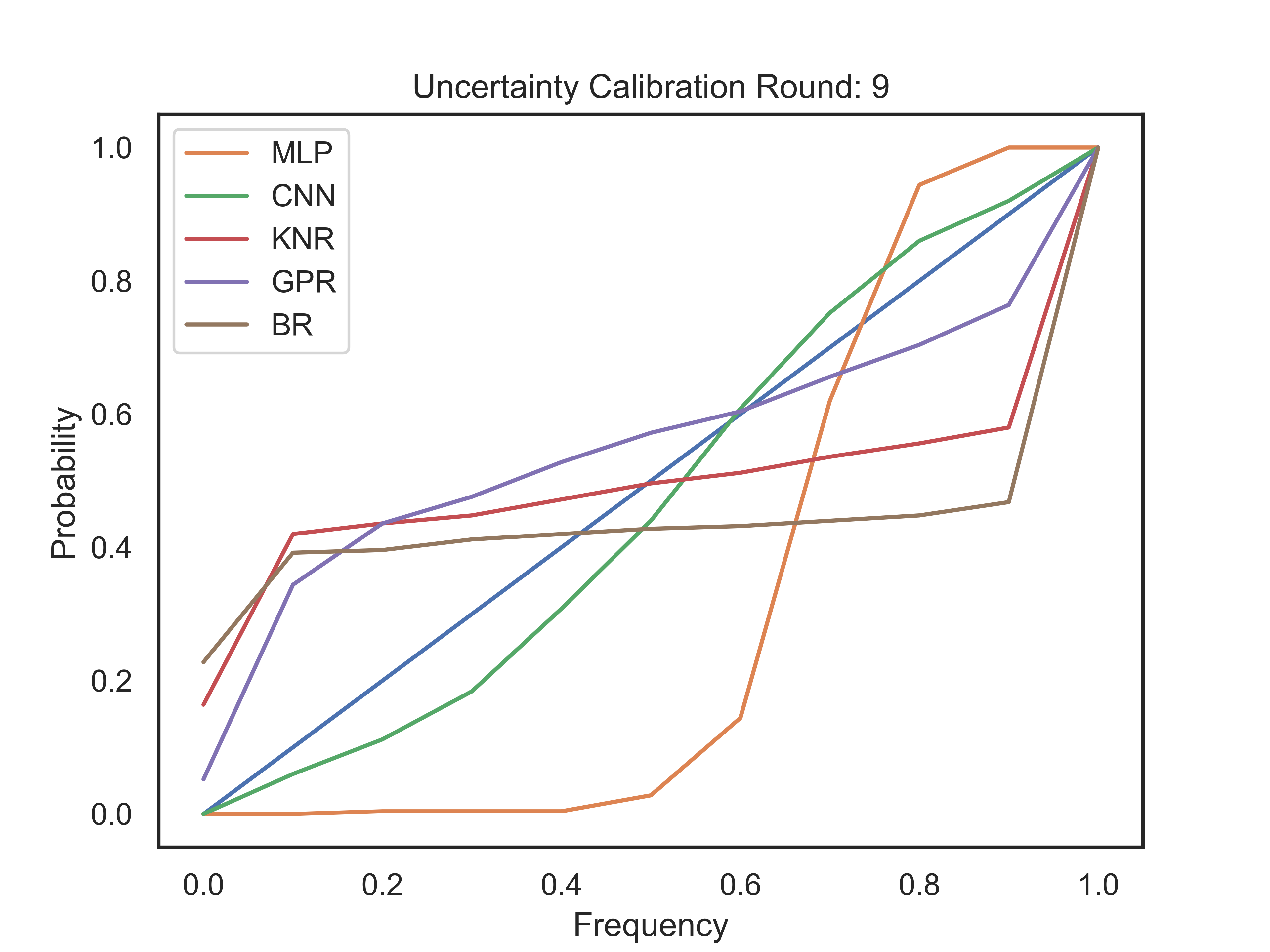}
    \end{subfigure}
    \hfill
    \begin{subfigure}[t]{0.32\textwidth}
        \centering
        \includegraphics[width=\linewidth]{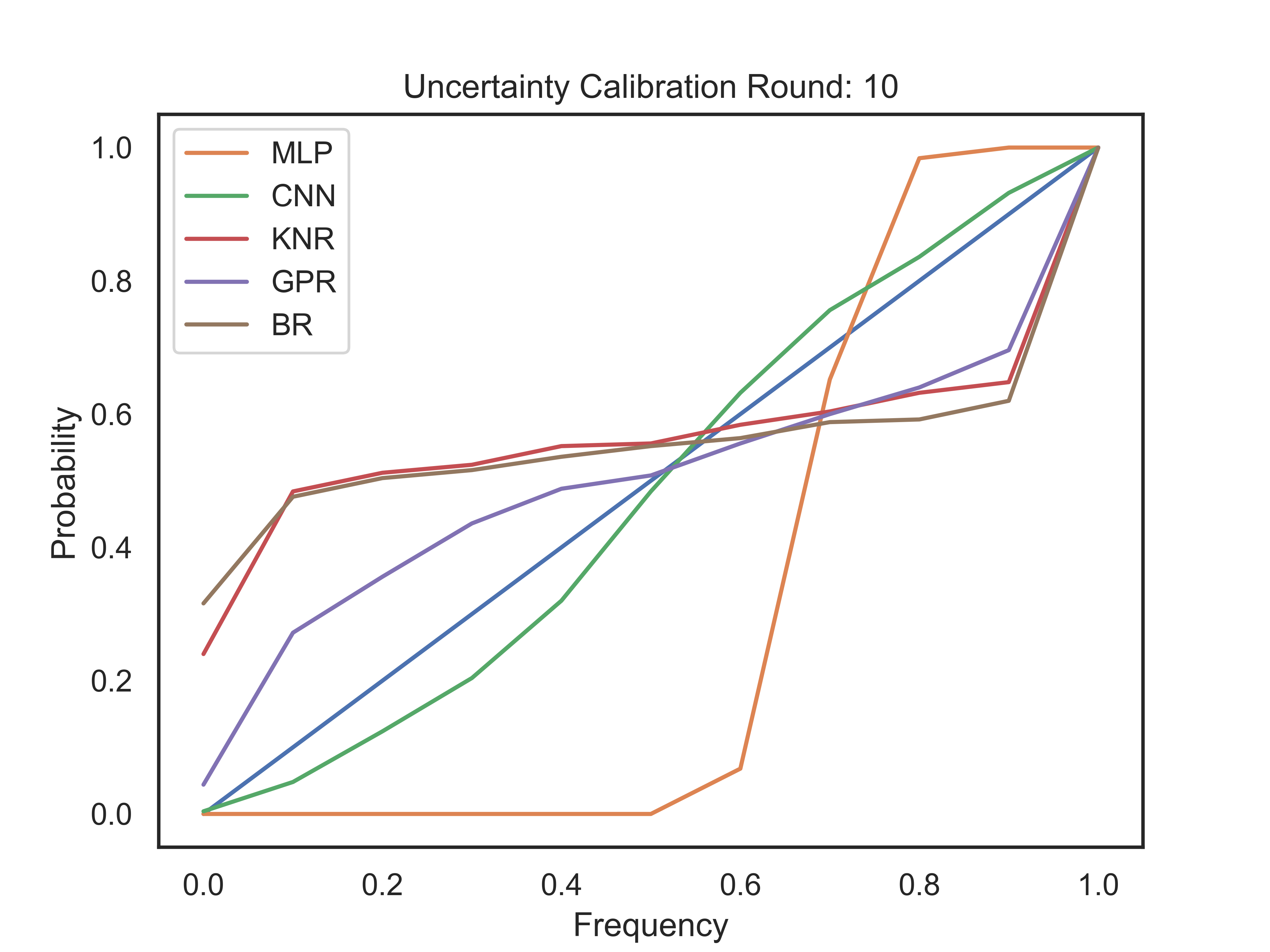}
    \end{subfigure}
    \hfill
    \caption{Comparison of the Uncertainty Calibration across different rounds. We can see that the ensemble of CNNs are better calibrated than the other surrogate models across the vast majority of the rounds.}
    \label{fig:ablation}
\end{figure*}

\end{document}